\documentclass[letterpaper, 10 pt, conference]{ieeeconf}  % Comment this line out if you need a4paper

\IEEEoverridecommandlockouts                              % This command is only needed if 
\usepackage{graphics} % for pdf, bitmapped graphics files
\usepackage{epsfig} % for postscript graphics files
\usepackage{amsmath} % assumes amsmath package installed
\usepackage{amssymb}  % assumes amsmath package installed
\usepackage{bm}  % assumes amsmath package installed
\usepackage{graphicx}
\usepackage{color}
\usepackage{mathtools}
\usepackage{array}
\usepackage{paralist}
\usepackage{nicefrac}
\usepackage{mathdots}
\usepackage{sidecap}
\usepackage{soul}
\usepackage{subcaption}
\usepackage[font=small,labelfont=small]{caption}
\usepackage{hyperref}
\newcommand{\etal}{\MakeLowercase{\textit{et al.}}}
\newcommand{\trsp}{\mathsf{T}}
\newcommand{\ty}[1]{{\scriptscriptstyle{\mathcal{#1}}}}
\DeclareMathOperator*{\argmax}{argmax} % thin space, limits underneath in displays

\definecolor{mygray}{RGB}{240,240,240}
\definecolor{rr}{rgb}{.8,0,0}
\definecolor{gg}{rgb}{0,.7,0}
\definecolor{bb}{rgb}{0,0,.8}
\definecolor{rb}{rgb}{.4,0,.4}

\title{\LARGE \bf
Interactive Trajectory Adaptation through \\Force-guided Bayesian Optimization
}

\author{Leonel Rozo$^{1}$% <-this % stops a space
	\thanks{$^{1}$ Bosch Center for Artificial Intelligence, Robert-Bosch-Campus 1, 71272 Renningen, Germany. {\tt\scriptsize leonel.rozo@de.bosch.com}}% <-this % stops a space
}

\begin{document}

\maketitle
\thispagestyle{empty}
\pagestyle{empty}

%%%%%%%%%%%%%%%%%%%%%%%%%%%%%%%%%%%%%%%%%%%%%%%%%%%%%%%%%%%%%%%%%%%%%%%%%%%%%%%%
%%%%%%%%%%%%%%%%%%%%%%%%%%%%%%%%%%%%%%%%%%%%%%%%%%%%%%%%%%%%%%%%%%%%%%%%%%%%%%%%
\begin{abstract}
Flexible manufacturing processes demand robots to easily adapt to changes in the environment and interact with humans. In such dynamic scenarios, robotic tasks may be programmed through learning-from-demonstration approaches, where a nominal plan of the task is learned by the robot. However, the learned plan may need to be adapted in order to fulfill additional requirements or overcome unexpected environment changes. When the required adaptation occurs at the end-effector trajectory level, a human operator may want to intuitively show the robot the desired changes by physically interacting with it. In this scenario, the robot needs to understand the human intended changes from noisy haptic data, quickly adapt accordingly and execute the nominal task plan when no further adaptation is needed. This paper addresses the aforementioned challenges by leveraging LfD and Bayesian optimization to endow the robot with data-efficient adaptation capabilities. Our approach exploits the sensed interaction forces to guide the robot adaptation, and speeds up the optimization process by defining local search spaces extracted from the learned task model. We show how our framework quickly adapts the learned spatial-temporal patterns of the task, leading to deformed trajectory distributions that are consistent with the nominal plan and the changes introduced by the human.  
\end{abstract}

%%%%%%%%%%%%%%%%%%%%%%%%%%%%%%%%%%%%%%%%%%%%%%%%%%%%%%%%%%%%%%%%%%%%%%%%%%%%%%%%
\section{INTRODUCTION}
The next generation of small and medium-sized enterprises (SMEs) is envisioned as places where humans and robots will fluently collaborate in a plethora of manufacturing processes. In current SMEs, the requirements of such processes may rapidly vary or a whole process could even change, requiring fast and efficient adaptation capabilities of workers. These conditions are clearly not a problem for human operators, as we exhibit extraordinary adaptation skills when faced with uncertain and highly-dynamic environments. However, when it comes to introduce robotic coworkers in SMEs, the required adaptation features for SMEs processes are still under development.  

Humans working in SMEs daily collaborate with each other in a large diversity of scenarios, showing complex adaptation capabilities that involve role specialization, motion re-planning, and role switching~\cite{ReedPeshkin:PhysicalHRI08}. Concerning human-robot collaboration (HRC), adaptation has been addressed either on one of the agents (i.e. leader-follower approaches), or on both of them (i.e. mutual adaptation)~\cite{Nikolaidis17:Models4HRC}. In both cases, robot adaptation is imperative to account for environment uncertainties, to respond to unseen task conditions, and to handle varying human preferences. In the latter case, it is difficult to build an accurate model of the human actions and internal state as it demands an understanding of human behavior when interacting with robotic systems. However, by exploiting the information obtained from several human-robot interactions and the human knowledge about the task, the aforementioned modeling and adaptation problems may be partly alleviated.  

When collaboration involves physical interaction among humans, haptic cues are often exploited to communicate intentions or negotiate roles~\cite{ReedPeshkin:PhysicalHRI08}. HRC settings may similarly exploit this information to, for example, guide the robot adaptation (i.e. exploration) when new actions are required for a new task situation. Argall~\etal~\cite{Argall10:TactileCorrection} used tactile feedback to convey human corrections for a policy that was initially learned from demonstrations. These tactile corrections were translated into incremental shifts for the robot pose, which were then used to either improve a current model or learn a different solution for the task. Human-guided exploration using force data was also employed to learn haptic affordances in robotic manipulation in~\cite{Chu16:HapticAffordance,Chu17:GuidedExploration}. The robot action search space was initialized from demonstrations, while the exploration was determined by the human guidance. The foregoing works provided the human with full control to govern the robot exploration, meaning that the robot was nearly behaving as a pure follower when human guidance took place. 

\begin{figure}[!tbp]
	\includegraphics[width=.22\textwidth]{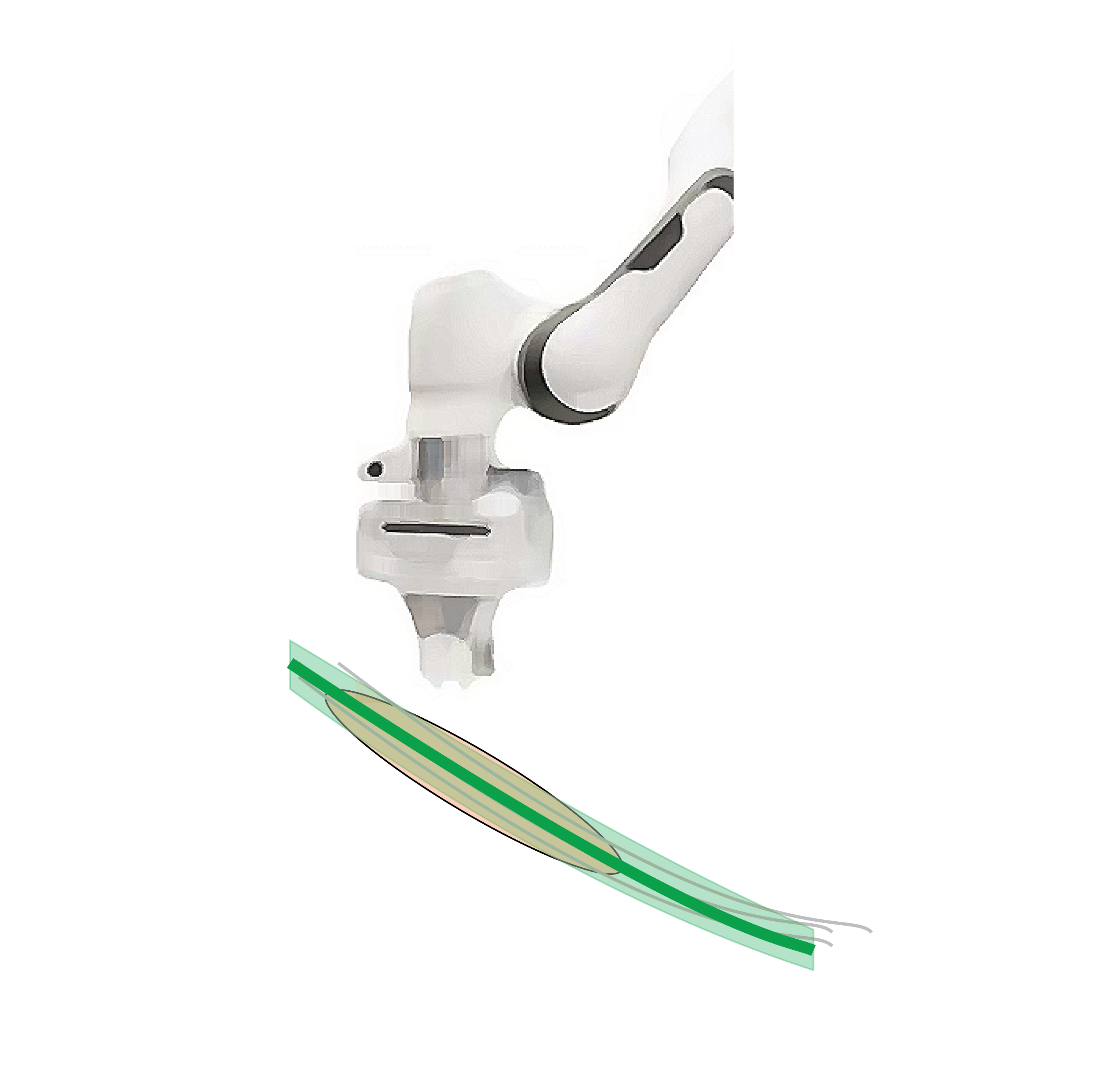}
	\includegraphics[width=.22\textwidth]{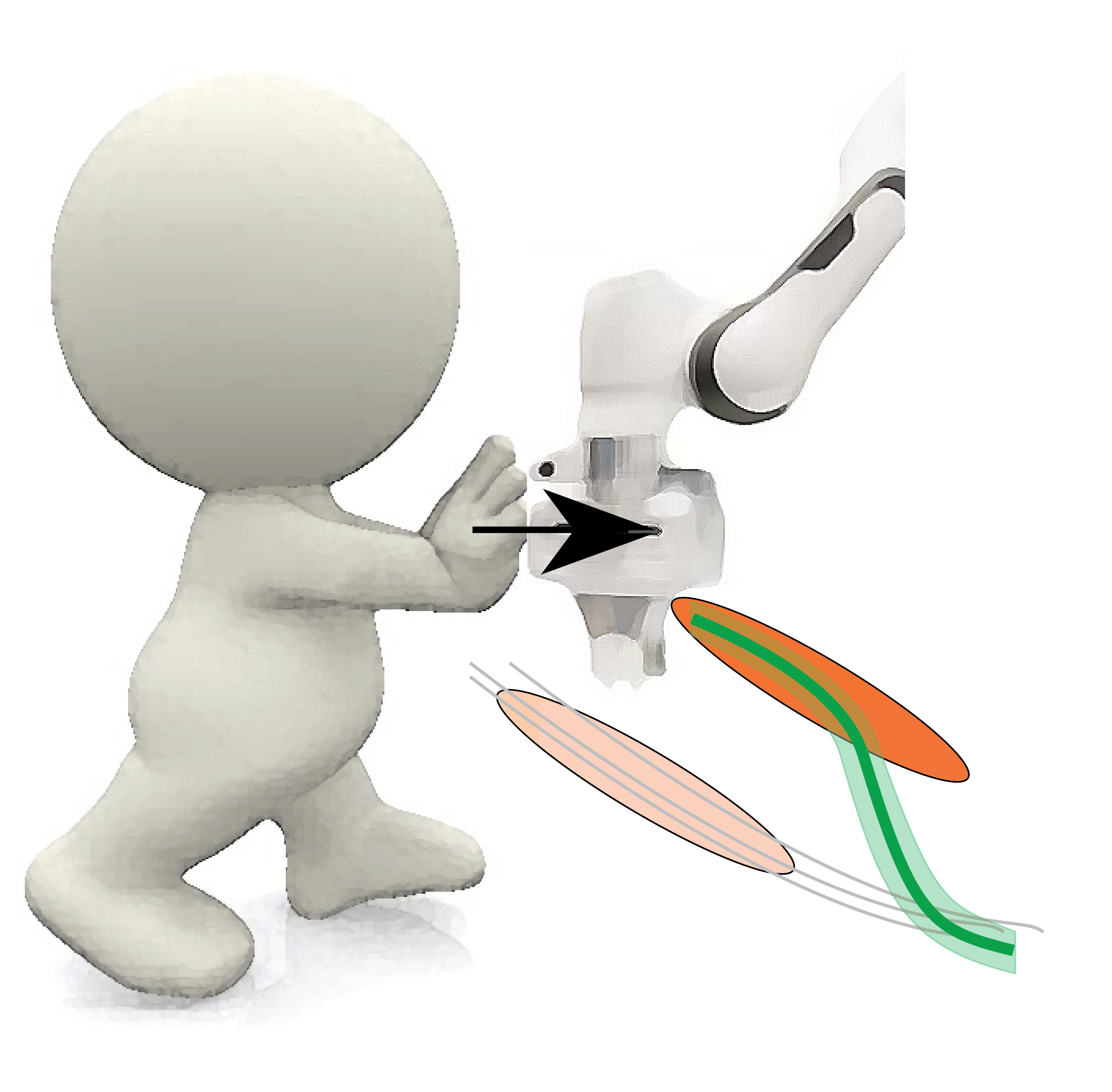}
	\caption{Illustration of the nominal plan execution learned from human demonstrations (\textit{left}), and the interactive trajectory adaptation phase (\textit{right}). When no adaptation is needed, the robot executes the task by following a reference trajectory distribution (green solid line depicting the mean and shared area representing the variance) extracted from the nominal plan (light orange ellipse). A human operator can introduce local trajectory deformations by physically interacting with the robot, which adapts the nominal task plan (dark orange ellipse) so that disagreement forces are minimized.}
	\label{Fig:Illustration}
\end{figure}

Robots can adapt more autonomously if human guidance provides only partial information about the possible exploration directions. Schroecker~\etal~\cite{Schroecker16:DirectingPolicySearch} proposed to use human demonstrations to provide the robot with salient (i.e. relevant) points of a trajectory while the robot autonomously learned the shape of the in-between trajectory. These points also served as online corrections to guide a reward-dependent robot exploration. However, handcrafting reward functions is tedious, error-prone, and becomes significantly more complex when a robot needs to collaborate with a human partner and adapt on-the-fly to her preferences.  

A potential alternative solution to endow robots with adaptation capabilities without explicitly defining an objective function is Bayesian optimization (BayesOpt)~\cite{Shahriari16:BayesOpt}. BayesOpt is a gradient-free global optimization of black-box objective functions that are expensive to evaluate and often multimodal. These characteristics properly fit the basic requirements of robots that need to adapt to new task conditions (i.e. expensive trials, unknown or hard-to-model objective function and data efficiency). BayesOpt has recently gained interest in different robotic applications, such as behavior adaptation for damaged robots~\cite{Cully15:RobotsAsAnimal}, automatic controller tuning for balancing~\cite{Marco16:LQRbayesOpt}, locomotion~\cite{Antonova17:DeepKernelsBO}, and interaction tasks~\cite{Drieb17:ConstBayesOptForceTask}, and in physical HRC~\cite{Ghadirzadeh16:RL4pHRI,Kupcsik15:HandoverBayes}. BayesOpt was used in~\cite{Ghadirzadeh16:RL4pHRI} to select the collaborative actions that minimize a Q-function in a model-based reinforcement learning approach. Interaction forces and positional data, representing the robot state, were exploited to determine the exploration actions. In~\cite{Kupcsik15:HandoverBayes}, the authors employed BayesOpt to approximate an unknown reward function for a contextual policy search aimed at learning a handover skill by interacting with a human. However, both~\cite{Ghadirzadeh16:RL4pHRI,Kupcsik15:HandoverBayes} did not exploit the fact that the robot may learn an initial policy from demonstrations, which may speed up the learning process and reduce explorations.     

Inspired by the insights on haptic communication~\cite{ReedPeshkin:PhysicalHRI08} and robot adaptation requirements in HRC~\cite{Nikolaidis17:Models4HRC}, we propose to exploit BayesOpt to adapt the robot nominal task plan through physical human guidance. Specifically, our framework learns a nominal plan of the task from human demonstrations using a hidden semi-Markov model~\cite{Yu:HSMM10,Rozo16Front:ADHSMM}. This model is later used to compute a smooth reference trajectory distribution representing the sensorimotor patterns of the task, as explained in Section~\ref{sec:Learning}. The nominal plan allows the robot to take a leading role during the task execution under normal conditions. A human operator can physically interact with the robot to indicate, through force-based cues, a desired adaptation for a new task situation, as illustrated in Fig.~\ref{Fig:Illustration}. BayesOpt is used to carry out a local adaptation of the nominal model so that the interaction forces (measuring the human-robot disagreement) are minimized (see Section~\ref{sec:BayesOpt}). 

A similar work on physically interactive trajectory deformations used an analytical smooth family of trajectories to find the local spatial deformations as a function of the applied force~\cite{Losey18:interactiveTrajDeformation}. The analytical formulation allowed to use gradient-based optimization to find the parameters of the deformed trajectory. In contrast, our approach allows for both spatial and temporal adaptation of the nominal task plan, and provides data-efficient adaptation by confining BayesOpt to carry out local searches at the level of the model states distribution, which significantly reduces the parameter space dimensionality. Moreover, the search space is automatically defined from the learning model, sharing some similarities with safe BayesOpt approaches~\cite{Berkenkamp16:safeBO}. The proposed framework is evaluated in Section~\ref{sec:Results} for different instances of a simulated 2D pick-and-place task, where data-efficient trajectory adaptations exploiting force-based guidance are successfully reported.

%%%%%%%%%%%%%%%%%%%%%%%%%%%%%%%%%%%%%%%%%%%%%%%%%%%%%%%%%%%%%%%%%%%%%%%%%%%%%%%%
\section{LEARNING A NOMINAL TASK PLAN}
\label{sec:Learning}
Robots can take advantage of learning-from-demonstration (LfD) approaches to learn to execute a task or collaborate with a human partner. Human demonstrations can be encoded by a probabilistic model that represents the nominal plan of the task for the robot. This model can subsequently be used to generate the desired robot movements as a function of the state of the human partner and the environment. To do so, we use a hidden semi-Markov model (HSMM)~\cite{Yu:HSMM10} to represent the nominal task plan of the robot. This model and its classical formulation (HMM) have been successfully exploited to learn manipulation skills with force sensing~\cite{RozoEtAl11:HMMGMRa,RozoEtAl13:ForceLfD}, in semi-autonomous teleoperation~\cite{Tanwani19:nonParamRobotLearn}, and in HRC~\cite{MedinaEtAl:CollForce11}.  HSMM allows us to encapsulate not only the observed sensorimotor patterns but also the temporal structure of the task. This model is then combined with a trajectory generation approach that exploits task dynamic features to retrieve a smooth reference distribution of sensorimotor trajectories. This is later used by the robot to both execute the desired task and monitor deviations that indicate an adaptation process. Both the learning model and retrieval of sensorimotor trajectories are described next.

\subsection{Hidden semi-Markov model}
\label{subsec:HSMM}
A $K$-states hidden Markov model (HMM) is characterized by an initial state distribution $\{\pi_i\}_{i=1}^K$, a transition probability matrix $\{a_{ij}\}_{i,j=1}^K$, and an observation distribution for each state $i$ in the model, commonly represented by a Gaussian distribution $\mathcal{N}(\bm{\mu}_i, \bm{\Sigma}_i)$, with mean $\bm{\mu}_i$ and covariance matrix $\bm{\Sigma}_i$. In HMM, the self-transition probabilities $a_{i,i}$ only allow a crude implicit modeling of the state duration which follows a geometric distribution $\mathcal{P}_i(d) = a_{i,i}^{d-1} (1-a_{i,i})$, decreasing exponentially with time~\cite{Rabiner89:TutorialHMM}. Thus, HMM is not suitable to encode tasks where temporal patterns are relevant. 

Variable duration modeling techniques such as \emph{hidden semi-Markov model} (HSMM) extend standard HMMs by embedding temporal information of the underlying stochastic process. That is, while in HMM the underlying hidden process is assumed to be Markov, i.e., the probability of transitioning to the next state depends only on the current state, in HSMM the state process is assumed semi-Markov.
This means that a transition to the next state depends on the current state as well as on the elapsed time since the state was entered~\cite{YuKobayashi:efficientHSMM06}. Since the state duration is always positive, its distribution should preferably be modeled by a function preserving this property. Thus, we here follow the approach proposed in~\cite{Rozo16Front:ADHSMM} and use a univariate lognormal distribution $\mathcal{N}(\mu^\ty{D}_i,\sigma^\ty{D}_i)$ with mean $\mu^\ty{D}_i$ and associated variance $\sigma^\ty{D}_i$ to model the logarithm of the duration, which is equivalent to the use of a lognormal distribution to fit the duration data. Therefore, an HSMM is characterized by the parameters set $ {\bm{\Theta}} = \left\{ \{a_{ij}\}_{j=1}^K, \mu_i^\ty{D}, \sigma_i^\ty{D}, \pi_i, \bm{\mu}_i, \bm{\Sigma}_i \right\}_{i=1}^K$, which can be trained by an expectation-maximization procedure.

Once trained, an HSMM can be used to derive a nominal task plan in the form of a desired sequence of states ${\bm{s}_{1:T}=\{s_1,s_2,\ldots,s_T\}}$ for a given time horizon of length $T$ and a set of discrete states ${s_t\in\{1,\ldots,K\}}$. To do so, we exploit the definition of the \emph{forward variable} in HSMM to compute the probability to be in state $i$ at time step $t$ and observe the partial observation $\bm{\zeta}_{1:t}~=~ \{\bm{\zeta}_1,\bm{\zeta}_2,\ldots,\bm{\zeta}_t\}$, namely $\alpha_{t,i} \triangleq \mathcal{P}(s_t\!=\!i \,,\, \bm{\zeta}_{1:t} )$, which is recursively computed with (see~\cite{Rabiner89:TutorialHMM} for details)\footnote{Equation~\eqref{eq:alphahsmm} can be efficiently computed as in~\cite{Yu:HSMM10, YuKobayashi:efficientHSMM06}.}
\begin{align}
\begin{split}
\alpha_{t,i}
&= \sum\limits_{d=1}^{d^{\max}} \sum\limits_{j=1}^K %\min(d^{\max}\!\!,t\!-\!1)}	
\alpha_{t-d,j} \; a_{j,i} \; \mathcal{N}^\ty{D}_{d,i}
\hspace{-.2cm}\prod\limits_{s=t-d+1}^{t}\hspace{-.2cm} \mathcal{N}_{s,i} ,%\\
\label{eq:alphahsmm}
\end{split}\\
\text{where}\quad \mathcal{N}^\ty{D}_{d,i}
&= \mathcal{N}\big(\log(d) |\; \mu^\ty{D}_i,\sigma^\ty{D}_i\big)
\;\quad\text{and}\;\\
\mathcal{N}_{s,i} &= \mathcal{N}\big(\bm{\zeta}_s |\; \bm{\mu}_i,\bm{\Sigma}_i\big) .\nonumber
\end{align}
For $t\!<\!d^{\max}$, the initialization is given by
\begin{gather*}
\alpha_{1,i} = \pi_i \; \mathcal{N}^\ty{D}_{1,i} \; \mathcal{N}_{1,i} ,
\\
\alpha_{2,i} = \pi_i \; \mathcal{N}^\ty{D}_{2,i} \prod_{s=1}^2 \! \mathcal{N}_{s,i}
+\!
\sum_{j=1}^K 	
\alpha_{1,j} \; a_{j,i} \; \mathcal{N}^\ty{D}_{1,i} \mathcal{N}_{2,i} ,
\\
\alpha_{3,i} = \pi_i \; \mathcal{N}^\ty{D}_{3,i} \prod_{s=1}^3 \!\ \mathcal{N}_{s,i}
+\!
\sum\limits_{d=1}^{2} \sum_{j=1}^K	
\alpha_{3-d,j} \; a_{j,i} \; \mathcal{N}^\ty{D}_{d,i}
\hspace{-.2cm}\prod\limits_{s=4-d}^{3}\hspace{-.2cm} \mathcal{N}_{s,i} ,
\end{gather*}
etc., which corresponds to the update rule
\begin{equation}
\alpha_{t,i} = \pi_i \; \mathcal{N}^\ty{D}_{t,i} \prod_{s=1}^t \! \mathcal{N}_{s,i}
+\!
\sum\limits_{d=1}^{t\!-\!1}	\sum_{j=1}^K
\alpha_{t-d,j} \; a_{j,i} \; \mathcal{N}^\ty{D}_{d,i}
\hspace{-.3cm}\prod\limits_{s=t-d+1}^{t}\hspace{-.3cm} \mathcal{N}_{s,i} .
\end{equation}

Then, given a time horizon $\{1\,:\,T\}$ and the definition of the forward variable~\eqref{eq:alphahsmm}, we can obtain the most likely sequence of states from 
\begin{equation}
	s_t = \argmax_{i} \alpha_{t,i} .
\end{equation}
HSMM can be viewed as a model representing a high-level abstraction of the task, which encapsulates the observed sensorimotor and temporal patterns through the set observation, duration and transition probabilities. This model is here exploited to \emph{(1)} retrieve a smooth reference trajectory distribution to drive the robot motion (Section~\ref*{subsec:trajGMM}) and \emph{(2)} localize the force-guided adaptation (Section~\ref{subsec:BayesOpt}). 

\subsection{Trajectory generation using dynamic features}
\label{subsec:trajGMM}
In order to retrieve the reference trajectory distribution from HSMM, we resort to an approach that exploits both static and dynamic features of the observed data, encapsulated in the observation and duration probability distributions. In robotics, this provides a simple approach to synthesize smooth trajectories, which is achieved by coordinating the distributions of both static and dynamic features in the considered time series. This approach has rarely been exploited in robotics, at the exception of the works from~\cite{Rozo16Front:ADHSMM,SugiuraEtal:MotionLearningTrajHMM11} employing it to represent object manipulation movements or collaborative behaviors. We here take advantage of this approach for retrieving a smooth reference trajectory distribution that will drive the robot motion according to the nominal task plan encoded by HSMM.

Formally, let us define the state of the robot as ${\bm{\xi} \in \mathbb{R}^D}$, which can represent the robot end-effector pose, its joint configuration, or be composed of additional sensory information such as sensed Cartesian forces or joint torques. For sake of simplicity, we here present the retrieval of a reference distribution of trajectories of the robot end-effector position ${\bm{x} \in \mathbb{R}^D}$, with $D=3$. However, the approach can be straightforwardly applied to alternative robot state representations as explained at the end of this section. 

For encoding robot movements, Cartesian velocities $\dot{\bm{x}}$ and accelerations $\ddot{\bm{x}}$ can be used as dynamic features of the robot motion. By considering an Euler approximation, they are computed as
\begin{equation}
\dot{\bm{x}}_t = \frac{\bm{x}_{t+1}-\bm{x}_t}{\Delta t} ,
\ddot{\bm{x}}_t = \frac{\dot{\bm{x}}_{t+1}-\dot{\bm{x}}_t}{\Delta t} =
\frac{\bm{x}_{t+2}-2\bm{x}_{t+1}+\bm{x}_t}{\Delta t^2} .
\label{eq:trajGMM_dx_ddx}
\end{equation}
By using \eqref{eq:trajGMM_dx_ddx}, the observation vector $\bm{\zeta}_t$ introduced in Section \ref{subsec:HSMM} will be used to represent the concatenated position, velocity and acceleration vectors at time step $t$, as follows
\begin{equation}
\bm{\zeta}_t=\left[
\begin{matrix}\bm{x}_t \\ \dot{\bm{x}}_t \\ \ddot{\bm{x}}_t\end{matrix}\right]
= \left[\begin{matrix}\bm{I} & \bm{0} & \bm{0} & \\
-\frac{1}{\Delta t}\bm{I} & \frac{1}{\Delta t}\bm{I} & \bm{0} \\
\frac{1}{\Delta t^2}\bm{I} & -\frac{2}{\Delta t^2}\bm{I} & \frac{1}{\Delta t^2}\bm{I}
\end{matrix}\right]
\left[
\begin{matrix} \bm{x}_t \\ \bm{x}_{t+1} \\ \bm{x}_{t+2}
\end{matrix}\right] ,
\label{eq:Phi1D}
\end{equation}
where $\bm{I} \in \mathbb{R}^{D\times D}$ is the identity matrix and ${\Delta t}$ the sampling time. Note that the number of derivatives is set up to acceleration, but the results can be generalized to a higher or lower number of derivatives. Then, new variables $\bm{\zeta}$ and $\bm{x}$ are defined as large vectors by concatenating $\bm{\zeta}_t$ and $\bm{x}_t$ for all time steps, namely $\bm{\zeta}=\left[\begin{matrix}\bm{\zeta}^\trsp_1 \> \bm{\zeta}^\trsp_2 \> \hdots \> \bm{\zeta}^\trsp_T \end{matrix}\right]^\trsp$ and ${\bm{x}=\left[\begin{matrix}\bm{x}^\trsp_1 \> \bm{x}^\trsp_2 \> \hdots \> \bm{x}^\trsp_T \end{matrix}\right]^\trsp}$. Similarly to the matrix operator \eqref{eq:Phi1D} defined for a single time step, a large sparse matrix $\bm{\Phi}$ can be defined so that $\bm{\zeta}\!=\!\bm{\Phi}\bm{x}$, namely\footnote{Note that a similar operator is defined to handle border conditions, and that $\bm{\Phi}$ can be constructed through the use of Kronecker products.}
\begin{equation}
\overbrace{
	\left[\begin{matrix}
	\vdots \\
	\color{rr}\bm{x}_t \\ \color{rr}\dot{\bm{x}}_t \\ \color{rr}\ddot{\bm{x}}_t \\
	\color{bb}\bm{x}_{t+1} \\ \color{bb}\dot{\bm{x}}_{t+1} \\ \color{bb}\ddot{\bm{x}}_{t+1} \\
	\vdots
	\end{matrix}\right]}^{\bm{\zeta}}\!\!=\!\!
\overbrace{
	\left[\arraycolsep=2.5pt
	\begin{matrix}
	\ddots & \vdots & \vdots & \vdots & \iddots & \\
	\cdots & \color{rr}\bm{I} & \color{rr}\bm{0} & \color{rr}\bm{0} & \cdots & \\
	\cdots & \color{rr}-\frac{1}{\Delta t}\bm{I} & \color{rr}\frac{1}{\Delta t}\bm{I} & \color{rr}\bm{0} & \cdots &\\
	\cdots & \color{rr}\frac{1}{\Delta t^2}\bm{I} & \color{rr}-\frac{2}{\Delta t^2}\bm{I} &
	\color{rr}\frac{1}{\Delta t^2}\bm{I} & \cdots & \\
	& \cdots & \color{bb}\bm{I} & \color{bb}\bm{0} & \color{bb}\bm{0} & \cdots\\
	& \cdots & \color{bb}-\frac{1}{\Delta t}\bm{I} & \color{bb}\frac{1}{\Delta t}\bm{I} & \color{bb}\bm{0} & \cdots\\
	& \cdots & \color{bb}\frac{1}{\Delta t^2}\bm{I} & \color{bb}-\frac{2}{\Delta t^2}\bm{I} &
	\color{bb}\frac{1}{\Delta t^2}\bm{I} & \cdots \\
	& \iddots & \vdots & \vdots & \vdots & \ddots\\
	\end{matrix}\right]}^{\bm{\Phi}}\!\!
\overbrace{
	\left[
	\begin{matrix}
	\vdots \\
	\color{rr}\bm{x}_t \\ \color{rb}\bm{x}_{t+1} \\ \color{rb}\bm{x}_{t+2} \\ \color{bb}\bm{x}_{t+3} \\
	\vdots
	\end{matrix}\right]}^{\bm{x}}\!.
\label{eq:zeta}
\end{equation}

The state sequence ${\bm{s}_{1:T}=\{s_1,s_2,\ldots,s_T\}}$ representing the nominal task plan can be exploited here to retrieve a reference trajectory distribution used to drive the robot end-effector movements. To do so, we define the likelihood of a movement $\bm{\zeta}$ for a given sequence $\bm{s}$ as
\begin{equation}
\mathcal{P}(\bm{\zeta}|\bm{s}) = \prod_{t=1}^T \mathcal{N}(\bm{\zeta}_t|\bm{\mu}_{s_t},\bm{\Sigma}_{s_t}) ,
\end{equation}
where $\bm{\mu}_{s_t}$ and $\bm{\Sigma}_{s_t}$ are the mean and covariance matrix of state $s_t$ at time step $t$. This state $s_t$, as defined in~\ref{subsec:HSMM}, is obtained from the most likely sequence of states sampled from the HSMM for a given horizon $\{1\,:\;T\}$ using the forward variable~\eqref{eq:alphahsmm} (note that we drop the subscript on $\bm{s}$ from now on). This product
can be rewritten as% the conditional distribution
\begin{gather}
\mathcal{P}(\bm{\zeta}|\bm{s}) = \mathcal{N}(\bm{\zeta}|\bm{\mu}_{\bm{s}},\bm{\Sigma}_{\bm{s}}) ,
\label{eq:Pzeta}
\\
\mathrm{with}\quad
\bm{\mu}_{\bm{s}} \!=\! \left[\begin{matrix}\bm{\mu}_{s_1} \\ \bm{\mu}_{s_2} \\ \vdots \\ \bm{\mu}_{s_T} \end{matrix}\right]
\:\text{and}\quad
\bm{\Sigma}_{\bm{s}} \!=\! \left[\begin{matrix}\bm{\Sigma}_{s_1} & \bm{0} & \cdots & \bm{0} \\
\bm{0} & \bm{\Sigma}_{s_2} & \cdots & \bm{0} \\
\vdots & \vdots & \ddots & \vdots \\
\bm{0} & \bm{0} & \cdots & \bm{\Sigma}_{s_T} \end{matrix}\right] .\nonumber
\end{gather}

By using the relation $\bm{\zeta}\!=\!\bm{\Phi}\bm{x}$, we then seek for a trajectory $\bm{x}$ maximizing the logarithm of \eqref{eq:Pzeta}, namely
\begin{equation}
\hat{\bm{x}} = \arg\max_{\bm{x}} \; \log\mathcal{P}(\bm{\Phi}\bm{x} \;|\;\bm{s}) .
\label{eq:MaxTrajGMM}
\end{equation}

The part of $\log\mathcal{P}(\bm{\Phi}\bm{x} \, | \, \bm{s})$ dependent on $\bm{x}$ takes the well-known quadratic error form ${c(\bm{x}) = 
{(\bm{\mu}_{\bm{s}}-\bm{\Phi}\bm{x})^\trsp \bm{\Sigma}_{\bm{s}}^{-1} (\bm{\mu}_{\bm{s}}-\bm{\Phi}\bm{x})}}$.
Then, a solution can be found by differentiating $c(\bm{x})$ and equating to $0$, providing the trajectory (in vector form)
\begin{equation}
\hat{\bm{x}} = \left(\bm{\Phi}^\trsp {\bm{\Sigma}_{\bm{s}}}^{-1} \bm{\Phi}\right)^{-1}
\bm{\Phi}^\trsp {\bm{\Sigma}_{\bm{s}}}^{-1} \bm{\mu}_{\bm{s}} ,
\label{eq:trajHMM}
\end{equation}
with the covariance error of the weighted least-squares estimate given by
\begin{equation}
\hat{\bm{\Sigma}}^{\bm{x}} = \sigma \left(\bm{\Phi}^\trsp {\bm{\Sigma}_{\bm{s}}}^{-1} \bm{\Phi}\right)^{-1} ,
\label{eq:trajHMM_cov}
\end{equation}
where $\sigma$ is a scale factor. Both \eqref{eq:trajHMM} and \eqref{eq:trajHMM_cov} describe a reference trajectory distribution $\mathcal{N}(\hat{\bm{x}}, \hat{\bm{\Sigma}}^{\bm{x}})$, which represents the nominal task plan learned from demonstrations. Note that the aforementioned equations can be computed efficiently with Cholesky and/or QR decompositions by exploiting the positive definite symmetric band structure of the matrices.

As we are interested in exploiting haptic cues to indicate desired adaptations to the learned nominal plan, we propose to augment the robot state as $\bm{\xi}\! = \! [\bm{x}^\trsp \bm{f}^\trsp]^\trsp$, where $\bm{f} \in \mathbb{R}^D$ represents the sensed Cartesian forces at the robot end-effector, leading to ${\bm{\xi} \in \mathbb{R}^{2D}}$ and $\bm{I} \in \mathbb{R}^{2D\times 2D}$. In this case, the dynamic features of the sensed forces are also considered ~\eqref{eq:zeta}. As a consequence, on the basis of~\eqref{eq:MaxTrajGMM}, the trajectory retrieval for this augmented state is formulated as
\begin{equation}
\hat{\bm{\xi}} = \arg\max_{\bm{\xi}} \; \log\mathcal{P}(\bm{\Phi}\bm{\xi} \;|\;\bm{s}) ,
\label{eq:MaxTrajGMMaugmented}
\end{equation}
whose solution has the same form as~\eqref{eq:trajHMM} and~\eqref{eq:trajHMM_cov}. This augmented state $\bm{\xi}$ allows us not only to retrieve a desired end-effector trajectory distribution $\mathcal{N}(\hat{\bm{x}}, \hat{\bm{\Sigma}}^{\bm{x}})$, but also a reference force distribution $\mathcal{N}(\hat{\bm{f}}, \hat{\bm{\Sigma}}^{\bm{f}})$. The latter can be employed to set a reference force profile to track during the execution of the task as well as to identify external forces indicating a task adaptation phase. In this work, the reference force distribution $\mathcal{N}(\hat{\bm{f}}, \hat{\bm{\Sigma}}^{\bm{f}})$ will be mainly used to detect adaptation phases from physical interactions.

%%%%%%%%%%%%%%%%%%%%%%%%%%%%%%%%%%%%%%%%%%%%%%%%%%%%%%%%%%%%%%%%%%%%%%%%%%%%%%%%
\section{TRAJECTORY ADAPTATION VIA FORCE-GUIDED BAYESIAN OPTIMIZATION}
\label{sec:BayesOpt}

Once a nominal plan has been learned, the robot is ready to carry out the task. However, during reproduction, new task requirements may arise (e.g., alternative motion trajectories, new locations of objects of interest), thus requiring the robot to adapt its nominal plan to the new situation. A way to indicate desired task adaptations is haptic communication. Here we allow a human operator to physically interact with the robot in order to show the required adaptation through force-based cues. Notice that in the case where no physical interaction is possible, artificial guidance forces could be computed from a virtual environment where a human commands a virtual proxy to indicate necessary adaptations. 

Let us assume that interaction forces convey information about the task adaptation required by a human. In other words, force-based cues provide information about an unobservable reward/objective function that the human is trying to optimize through the trajectory adaptation. As crafting reward/objective functions is significantly cumbersome and data-efficient adaptation is imperative when a robot interacts with a human, we propose to exploit BayesOpt to adapt the parameters of the nominal plan. We here provide a short introduction to BayesOpt and later explain how this is exploited into the force-guided robot adaptation.  

% 	- Background
\subsection{Bayesian optimization}
\label{subsec:BayesOpt}
In general terms, we are considering the problem of finding a global maximizer (or minimizer) of an unknown objective function $f$ 
\begin{equation}
	\bm{\theta}^* = \argmax_{\bm{\theta} \in \mathcal{X}} f(\bm{\theta}) ,
\end{equation}
where $\mathcal{X} \subseteq \mathbb{R}^{D_{\mathcal{X}}}$ is some design space of interest, with $D_{\mathcal{X}}$ being the dimensionality of the parameter space. Furthermore, we assume that the black-box function $f$ has no simple closed form, but can be evaluated at any arbitrary query point $\bm{\theta}$ in the domain. This evaluation produces noise-corrupted outputs $y \in \mathbb{R}$ such that ${\mathbb{E}[y | f(\bm{\theta})] = f(\bm{\theta})}$. In other words, we can only observe the function $f$ through unbiased noisy point-wise observations $y$. In this setting, we consider a sequential search algorithm which, at iteration $n$, selects a location $\bm{\theta}_{n+1}$ at which to query $f$ and observe $y_{n+1}$. After $N$ queries, the algorithm makes a final recommendation $\bm{\theta}_N$, which represents the algorithm's best estimate.

BayesOpt prescribes a prior belief over the possible objective functions and then sequentially refines this model as data are observed via Bayesian posterior updating. Equipped with this probabilistic model, BayesOpt can sequentially induce acquisition functions $\gamma_n : \mathcal{X} \mapsto \mathbb{R}$ that leverage the uncertainty in the posterior to guide the exploration. Intuitively, the acquisition function evaluates the utility of candidate points for the next evaluation of $f$; therefore, $\bm{\theta}_{n+1}$ is selected by maximizing $\gamma_n$, where the index $n$ indicates the implicit dependence on the currently available data.

A common way to model the prior and posterior for $f$ is by using a Gaussian Process $f(\bm{\theta}) \sim \mathcal{GP}(\mu(\bm{\theta}), k(\bm{\theta}_i, \bm{\theta}_j))$ with mean function $\mu :  \mathcal{X} \mapsto \mathbb{R}$ and positive-definite kernel (or covariance function) $k : \mathcal{X} \times \mathcal{X} \mapsto \mathbb{R}$. Let $\mathcal{D}_n = \{(\bm{\theta}_i, y_i)\}_{i=1}^n$ denote the set of observations and $\tilde{\bm{\theta}}$ represent an arbitrary test point. The random variable $f(\tilde{\bm{\theta}})$ conditioned on observations $\mathcal{D}_n$ is also normally distributed with the following posterior mean and variance functions:
\begin{align}
	\mu_n(\bm{\theta}) &= \mu(\tilde{\bm{\theta}}) + \bm{k}(\tilde{\bm{\theta}})^\trsp (\bm{K} + \sigma^2 \bm{I})^{-1}(\bm{y} - \bm{\mu}(\bm{\theta})), \\
	\sigma_n^2(\bm{\theta}) &= k(\tilde{\bm{\theta}},\tilde{\bm{\theta}}) - \bm{k}(\tilde{\bm{\theta}})^\trsp (\bm{K} + \sigma^2 \bm{I})^{-1} \bm{k}(\tilde{\bm{\theta}}), 
\end{align}
where $\bm{y}$ is the observed outputs vector, $\bm{k}(\tilde{\bm{\theta}})$ is a vector of covariance terms between $\tilde{\bm{\theta}}$ and $\bm{\theta}_{1:n}$, and $\bm{K}$ is the covariance matrix for all the pairs $\bm{\theta}_i$ and $\bm{\theta}_j$. The posterior mean and variance evaluated at any point $\tilde{\bm{\theta}}$ represent the model prediction and uncertainty, respectively, in the objective function at the point $\tilde{\bm{\theta}}$. These posterior functions are exploited to select the next query point $\bm{\theta}_{n+1}$ by means of an acquisition function. 

An acquisition function performs a trade-off between exploitation (e.g. selecting the point with the highest posterior mean) and exploration (e.g. selecting the point with the highest posterior variance) using the information given by the posterior functions. Here we use expected improvement (EI), which incorporates the amount of improvement upon $\tau$, and can be analytically computed as follows 
\begin{align}
	\gamma_{EI}(\bm{\theta}; \mathcal{D}_n) & = (\mu_n(\bm{\theta}) - \tau) \Phi\left( \frac{\mu_n(\bm{\theta}) - \tau}{\sigma_n(\bm{\theta})} \right) \nonumber \\
	&+ \sigma_n(\bm{\theta}) \phi\left( \frac{\mu_n(\bm{\theta}) - \tau}{\sigma_n(\bm{\theta})} \right) ,
	\label{eq:EI}
\end{align}
where $\Phi$ is the normal cumulative distribution function, $\phi$ represents the corresponding probability density function, and $\tau$ is the threshold improvement. Intuitively, EI selects the next parameter point where the expected improvement over $\tau$ is maximal (more details are given in~\cite{Shahriari16:BayesOpt,Mockus75:EI}).  

\subsection{Force-guided robot adaptation}
\label{subsec:RobotAdapt}
As mentioned previously, the human partner can physically interact with the robot to indicate, through force-based cues, a desired adaptation of the learned nominal plan. In order to let the robot negotiate its adaptation according to the human intention (which is noisily observed), we here exploit BayesOpt to carry out a local adaptation of the nominal model so that the interaction forces (measuring the human-robot disagreement) are minimized. As our learning model encapsulates both sensorimotor and duration patterns, spatiotemporal adaptations are possible. 

Formally, sensorimotor patterns and duration information are locally encoded as Gaussian distributions $\mathcal{N}(\bm{\mu}_i, \bm{\Sigma}_i)$ and $\mathcal{N}(\mu_i^{\mathcal{D}}, \sigma_i^{\mathcal{D}})$, as described in Section \ref{sec:Learning}. These distributions directly influence the robot task execution through~\eqref{eq:alphahsmm} and~\eqref{eq:MaxTrajGMM}. So, in order to locally adapt the robot trajectory, we define the vector of local model parameters as
\begin{equation}
\bm{\theta}_i = \left[
\begin{matrix}\bm{\mu}_i^{\mathcal{O}} \\ \mu_i^{\mathcal{D}} \end{matrix}\right] , 
\label{eq:thetaParams}
\end{equation}
where $\bm{\mu}_i^{\mathcal{O}}$ and $\mu_i^{\mathcal{D}}$ respectively represent the mean vector of motor commands and duration for state $i$.\footnote{Note that we omit covariance parameters for simplicity.} The vector of parameters $\bm{\theta}_i$ is thus optimized using BayesOpt to find the optimal $\bm{\theta}_{i}^*$ that minimizes the human-robot disagreement forces when a desired adaptation is triggered. Note that when no desired force profile is required for the task, the disagreement forces directly correspond to the noisy readings $\bm{f}^s$ of the force sensor mounted at the robot wrist. On the contrary, if a reference force distribution $\mathcal{N}(\hat{\bm{f}}, \hat{\bm{\Sigma}}^{\bm{f}})$ is given, the disagreement forces can be easily computed as the L-2 norm of the difference between the reference $\hat{\bm{f}}$ and the sensed forces $\bm{f}^s$. Therefore, in our case, the set of observations used to compute the posterior mean and variance functions is $\mathcal{D}_n = \{(\bm{\theta}_{i,j}, ||\hat{\bm{f}}-\bm{f}^s||_j)\}_{j=1}^n$.

The fact that the learning model encodes the nominal plan using a set of states (represented by Gaussian distributions) allows us to carry out a local search of the optimal parameters by identifying the state $i$ in which the robot is when a desired adaptation is triggered. To do so, we exploit the definition of the forward variable~\eqref{eq:alphahsmm} and choose the local adaptation parameters $\bm{\theta}_i$ as those corresponding to the state 
\begin{equation}
	i = \argmax_j \alpha_{t,j} .
	\label{eq:argmax_i}
\end{equation}
This reduces the dimensionality of the parameter space in contrast to a high-dimensional vector $\bm{\theta}$ composed of all the mean vectors of motor commands and durations of the learning model. Moreover, the domain $\mathcal{X}$ can be automatically extracted from human demonstrations, where both $\bm{\Sigma}_i^{\mathcal{O}}$ and $\sigma_i^{\mathcal{D}}$ specify local bounds in which BayesOpt is allowed to look for the optimal $\bm{\theta}_{i,N}$. For example, lower and upper bounds for the sensorimotor component of $\bm{\theta}_i$ may be defined as ${(\bm{\mu}_i^{\mathcal{O}}-2\bm{\sigma}_i^{\mathcal{O}}, \bm{\mu}_i^{\mathcal{O}}+2\bm{\sigma}_i^{\mathcal{O}})}$, where $\bm{\sigma}_i^{\mathcal{O}}$ is the variance vector in $\bm{\Sigma}_i^{\mathcal{O}}$. Interestingly, this automatic domain extraction shares similarities with the modified BayesOpt approach proposed in~\cite{Berkenkamp16:safeBO} for safe controllers optimization.

\begin{figure*}[!tbp]
	\includegraphics[width=.245\textwidth]{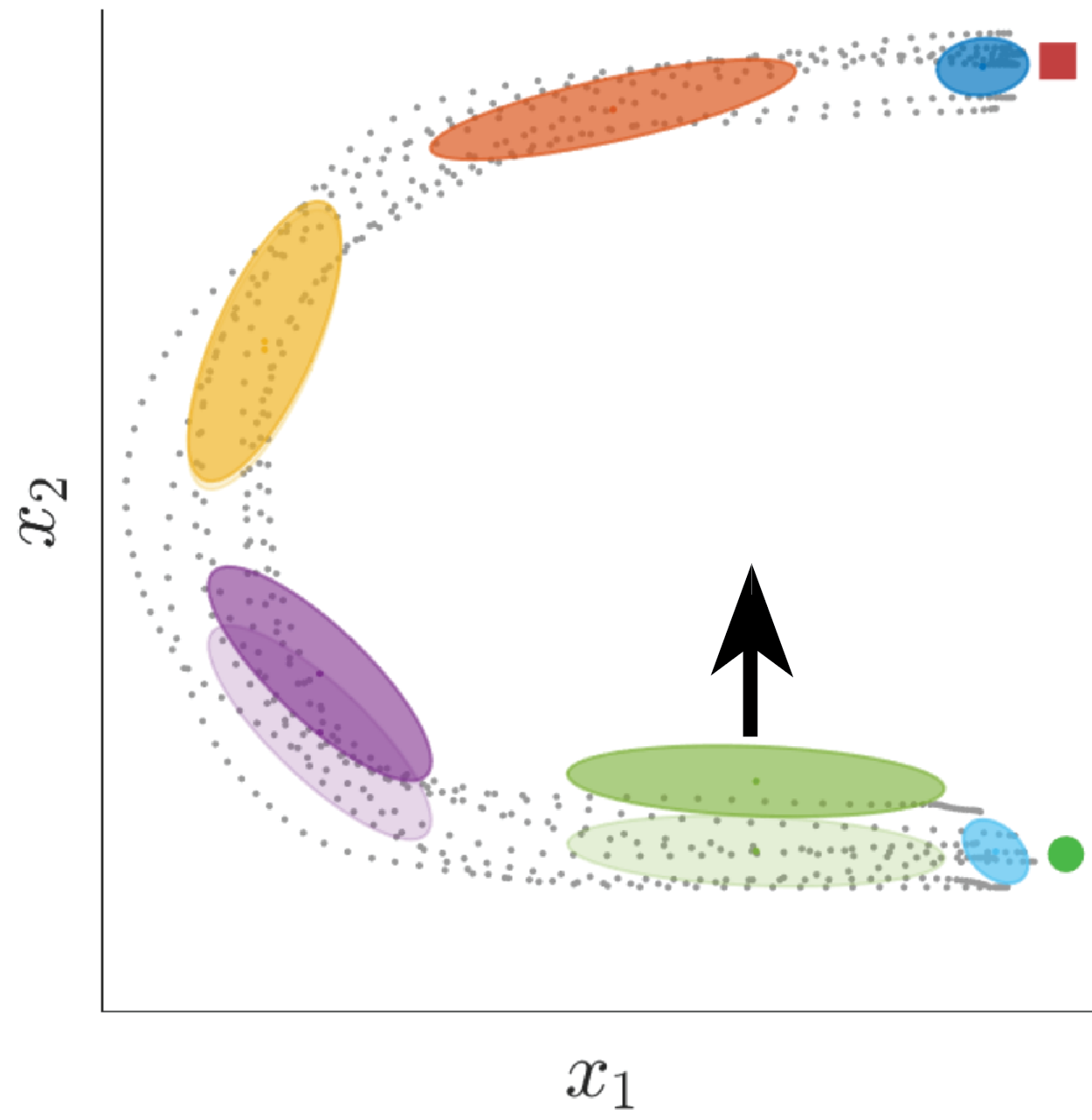}
	\includegraphics[width=.245\textwidth]{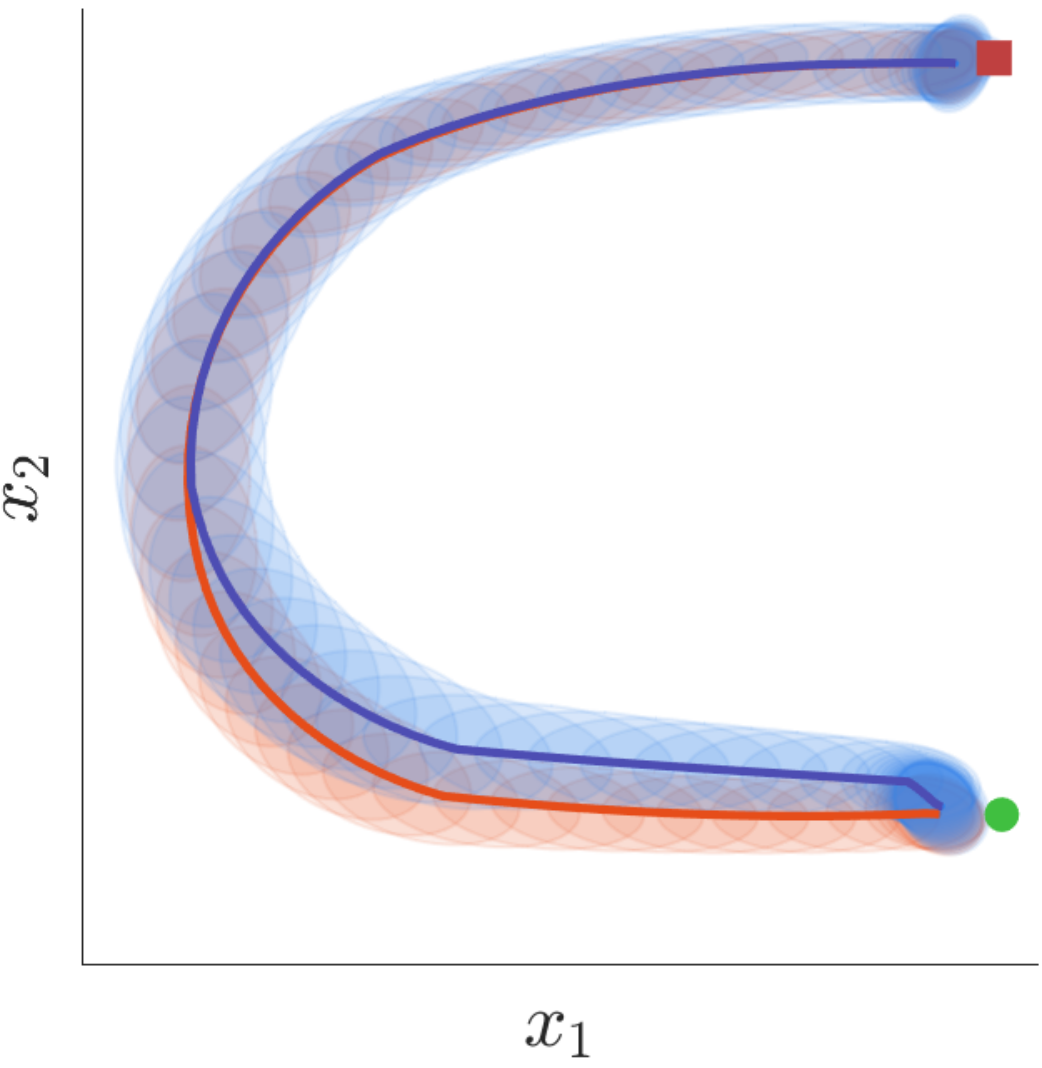}
	\includegraphics[width=.245\textwidth]{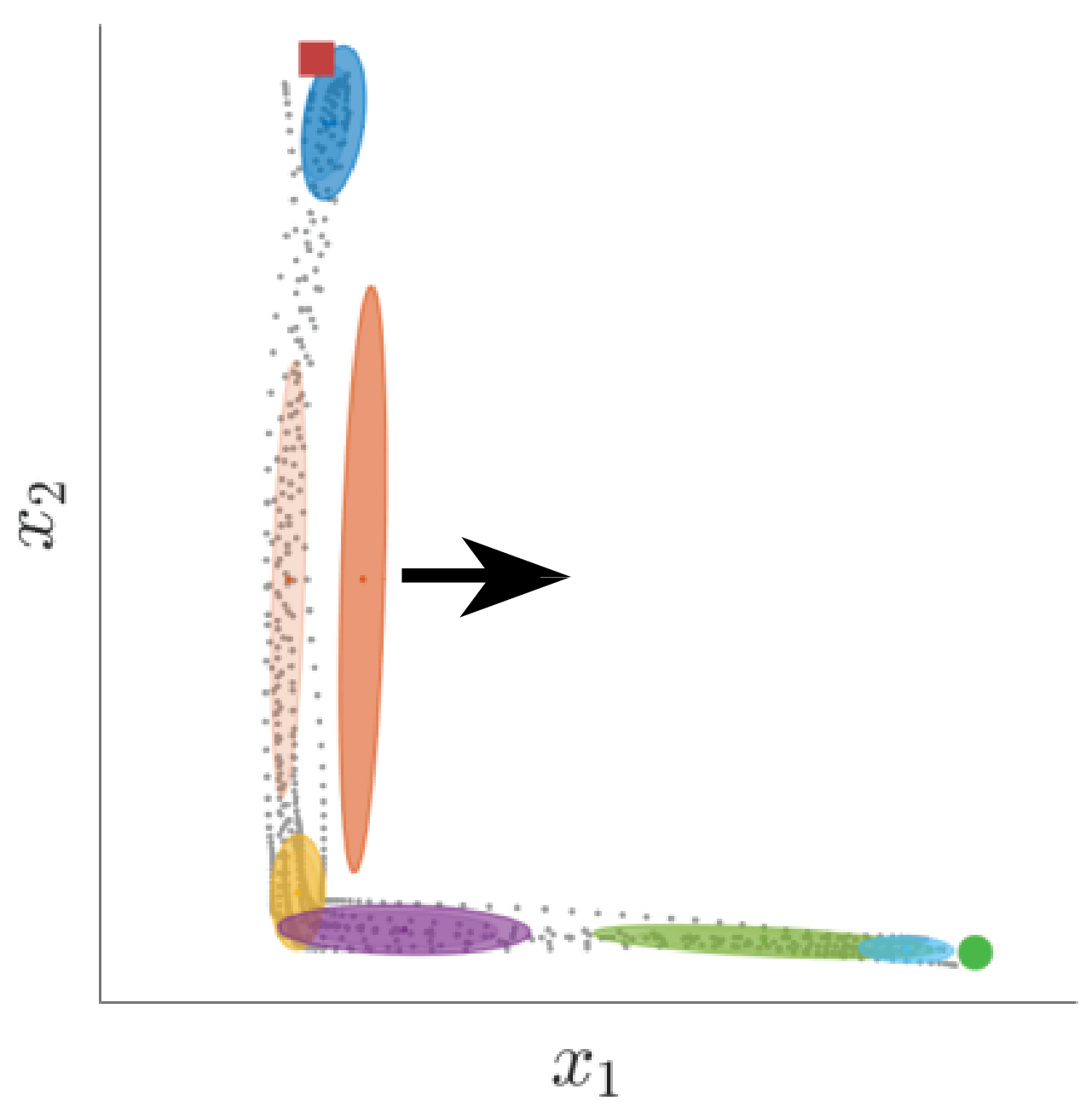}
	\includegraphics[width=.245\textwidth]{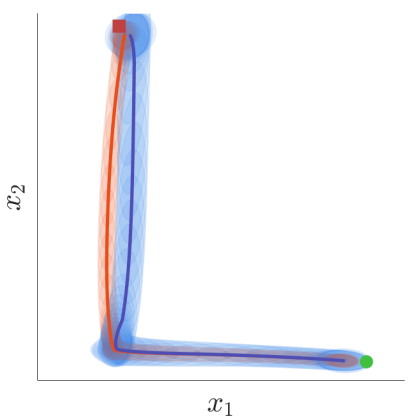}
	\caption{2D pick-and-place task (C- and L-shape trajectories). \emph{Left:} Demonstration data of 2D trajectories are displayed as gray points, while the nominal task plan encoded by an HSMM is shown as light-color ellipses. The adapted model is displayed using the same color format but with darker tones. The direction of the applied force is represented by the black arrows. \emph{Right:} the trajectory distributions obtained from nominal plan and force-guided BayesOpt adaptation are respectively displayed in red and blue. The solid line depicts the mean of the trajectory while the covariance is displayed as ellipses centered in the mean.}
	\label{Fig:NomAdapPlan}
\end{figure*}

\subsection{Online adapted trajectory generation}
When human intervention is detected, meaning that trajectory adaptation should take place, it is necessary to update the reference trajectory over the course of the task. To do so, every time our force-based local BayesOpt finds a set of optimal parameters $\bm{\theta}_{i,N}$, a new reference distribution of trajectories is generated by computing a new sequence of states $\bm{s}_{t:T_w}$ via~\eqref{eq:alphahsmm} for a time window of length $T_w$. This is later used to generate the new trajectory distribution through~\eqref{eq:trajHMM} and~\eqref{eq:trajHMM_cov}. Note that the specification of a time window assumes that the interactive trajectory adaptation occurs for relatively short time periods, meaning that the robot is expected to resume the execution of the nominal plan once the human operator does not trigger any adaptation. Moreover, the time window favors the computational cost of the whole adaptation process. 	

\begin{figure}[!tbp]	
	\includegraphics[width=.2\textwidth]{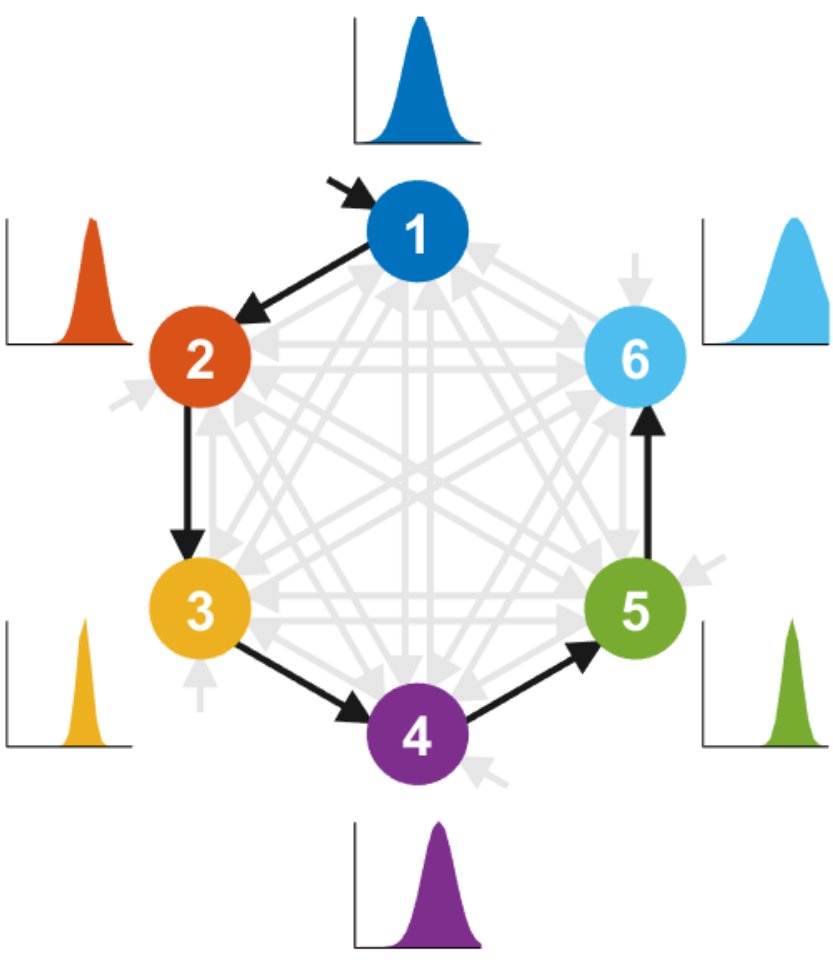}
	\hspace{0.3cm}
	\includegraphics[width=.24\textwidth]{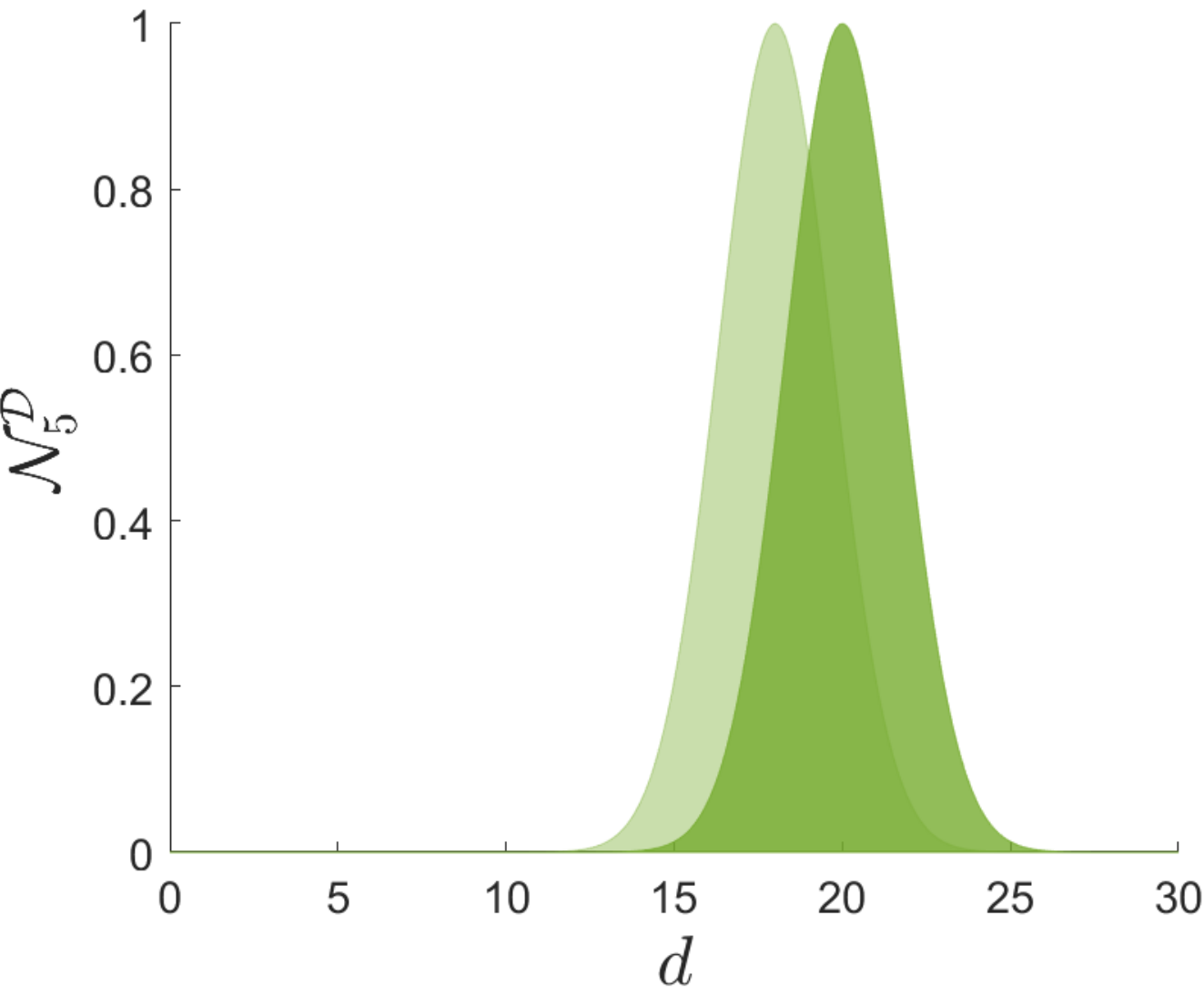}
	\caption{2D pick-and-place task (C-shape trajectories). \emph{Left:} HSMM transition graph and state duration probabilities as lognormal distributions, with the six states depicted with the same colors as the Gaussian distributions in Fig. \ref{Fig:NomAdapPlan}. \emph{Right:} Original and adapted duration probability distributions of the fifth HSMM state in encoding of C-shape trajectories shown in Fig. \ref{Fig:NomAdapPlan}.}
	\label{Fig:TransDurNominalPlan}
\end{figure}

\begin{figure}[!tbp]
	\centering
	\includegraphics[width=.4\textwidth]{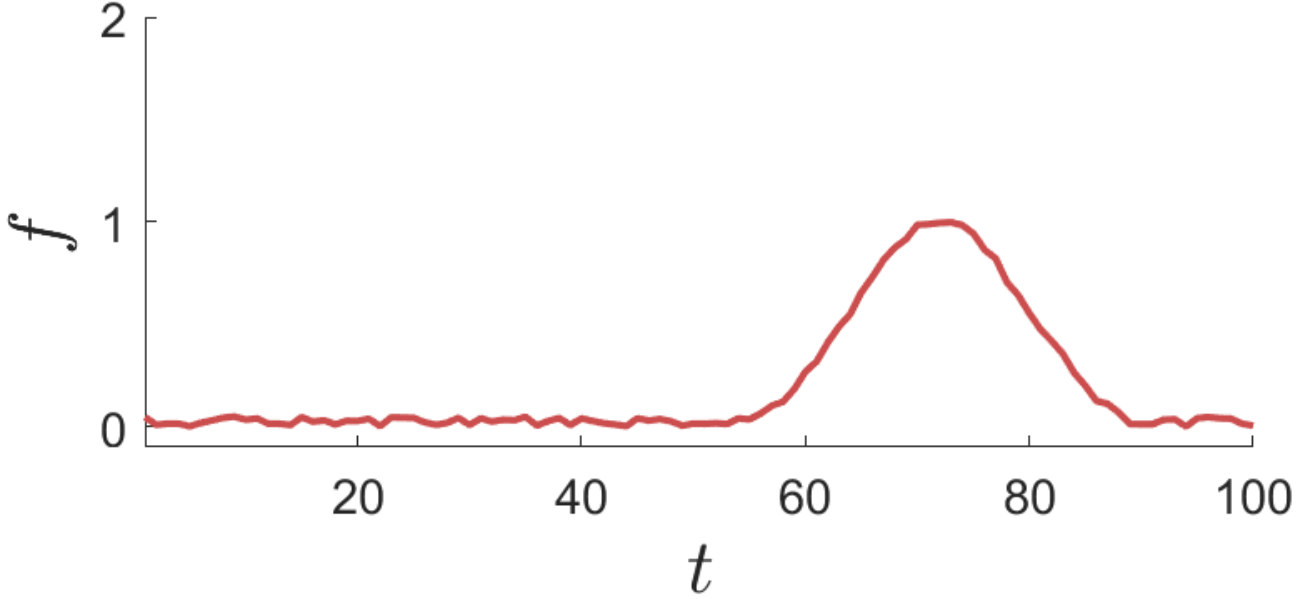}
	\caption{Force profile applied to the robot end-effector in order to introduce force-guided local trajectory adaptations.}
	\label{Fig:InteractionForce}
\end{figure}

%%%%%%%%%%%%%%%%%%%%%%%%%%%%%%%%%%%%%%%%%%%%%%%%%%%%%%%%%%%%%%%%%%%%%%%%%%%%%%%%
\section{EXPERIMENTS}
\label{sec:Results}

\subsection{Description}
The pick-and-place task is a standard setup where a robotic manipulator is required to reach for an object, grasp it, and consequently move it to a target location into its workspace, where the object is released. This scenario is highly relevant in SMEs as its main characteristics require range of common robot skills that are found in other types of tasks. For the experiments, we collected six synthetic demonstrations where both 2D Cartesian trajectories and sensed force profiles were generated while the robot followed C- and L-shape trajectories, as shown in Fig.~\ref{Fig:NomAdapPlan}. Note that in this particular task the robot does not require to apply a specific force while moving the object, which means that the recorded sensed forces are zero-mean. 
For both datasets, we trained a couple of 6-states HSMMs to learn the nominal plan of the task, which is mainly encapsulated by the set of normal distributions $\{ \mathcal{N}_{s,i}, \mathcal{N}^{\mathcal{D}}_{i}\}_{i=1}^K$ encoding local sensorimotor patterns and duration information.

During reproduction of the task, the robot end-effector movement is mainly driven by the reference trajectory distribution computed by~\eqref{eq:trajHMM} and~\eqref{eq:trajHMM_cov}. The detection of human intervention is carried out by monitoring the difference between the reference force distribution $\mathcal{N}(\hat{\bm{f}}_t, \hat{\bm{\Sigma}}^{\bm{f}}_t)$ and the sensed forces $\bm{f}^s_t$. A time window of five time steps was used to compute these disagreement forces. If human intervention is detected, a local adaptation process is triggered by running the forge-guided local search described in Section~\ref{subsec:RobotAdapt}. This local search is implemented by proposing new local parameters $\bm{\theta}_{i,t+1}$ at each time step $t$ according to the acquisition function~\eqref{eq:EI}. Once the optimal set of local parameters $\bm{\theta}_i^*$ has been found, the reference trajectory distribution is recomputed using the updated observation and duration probabilities of state $i$ with new means $\bm{\mu}_i^{\mathcal{O}^*}$ and $\mu_i^{\mathcal{D}^*}$. Note that this simple example for testing our trajectory adaptation approach resembles the robot negotiation for human intervention in the context of a mail delivery task~\cite{Yin18:HumanInRobotMotionLearning}.

\subsection{Learning and reproduction of nominal plan}
Figure~\ref{Fig:NomAdapPlan} shows the encoding of the nominal plan of the task extracted from the synthetic demonstrations for both datasets (i.e, C- and L-shape trajectories). The ellipses represent the HSMM observation probability distributions $\mathcal{N}_{s,i}$, where light tones correspond to the original nominal task plan. The reference trajectory distribution driving the end-effector motion during the execution of the nominal plan (i.e. no adaptation is triggered) is depicted by the red solid line (mean) and shared area (covariance). The transition graph and state duration probabilities $\mathcal{N}^{\mathcal{D}}_{i}$ for the HSMM encoding the C-shape trajectories are displayed in Fig.~\ref{Fig:TransDurNominalPlan}. As expected, the reference trajectory distribution retrieved by using the HSMM encoding properly encodes the sensorimotor patterns observed in the demonstrations.

\subsection{BayesOpt-based adaptation}
It is more interesting though to notice the effects of using force-guided BayesOpt. To do so, a simulated external force was applied to the robot end-effector during the execution of the nominal plan. The direction of the applied force is represented by the black arrows in Fig.~\ref{Fig:NomAdapPlan} while its profile is shown in Fig.~\ref{Fig:InteractionForce}. The first and third plot in Fig.~\ref{Fig:NomAdapPlan} show the adapted model with darker-tone ellipses, where some of the them fully overlap the ellipses depicting the original plan as the force-guided adaptation was triggered only for a specific time horizon (approximately from $t=55$ to $t=90$). As it can be observed, the purple and green ellipses -- in the C-shape trajectories model-- were vertically translated as an agreement with the external force. A similar effect was produced by the force-guided BayesOpt in the L-shape trajectories model where the orange Gaussian was horizontally moved. Additionally, our adaptation process also modified the duration probability distributions when necessary. Figure~\ref{Fig:TransDurNominalPlan}-\emph{right} shows how the duration probability distribution of the fifth HSMM state was adapted so that the robot stays longer in it.   

\begin{figure}[!tbp]
	\centering
	\includegraphics[width=.45\textwidth]{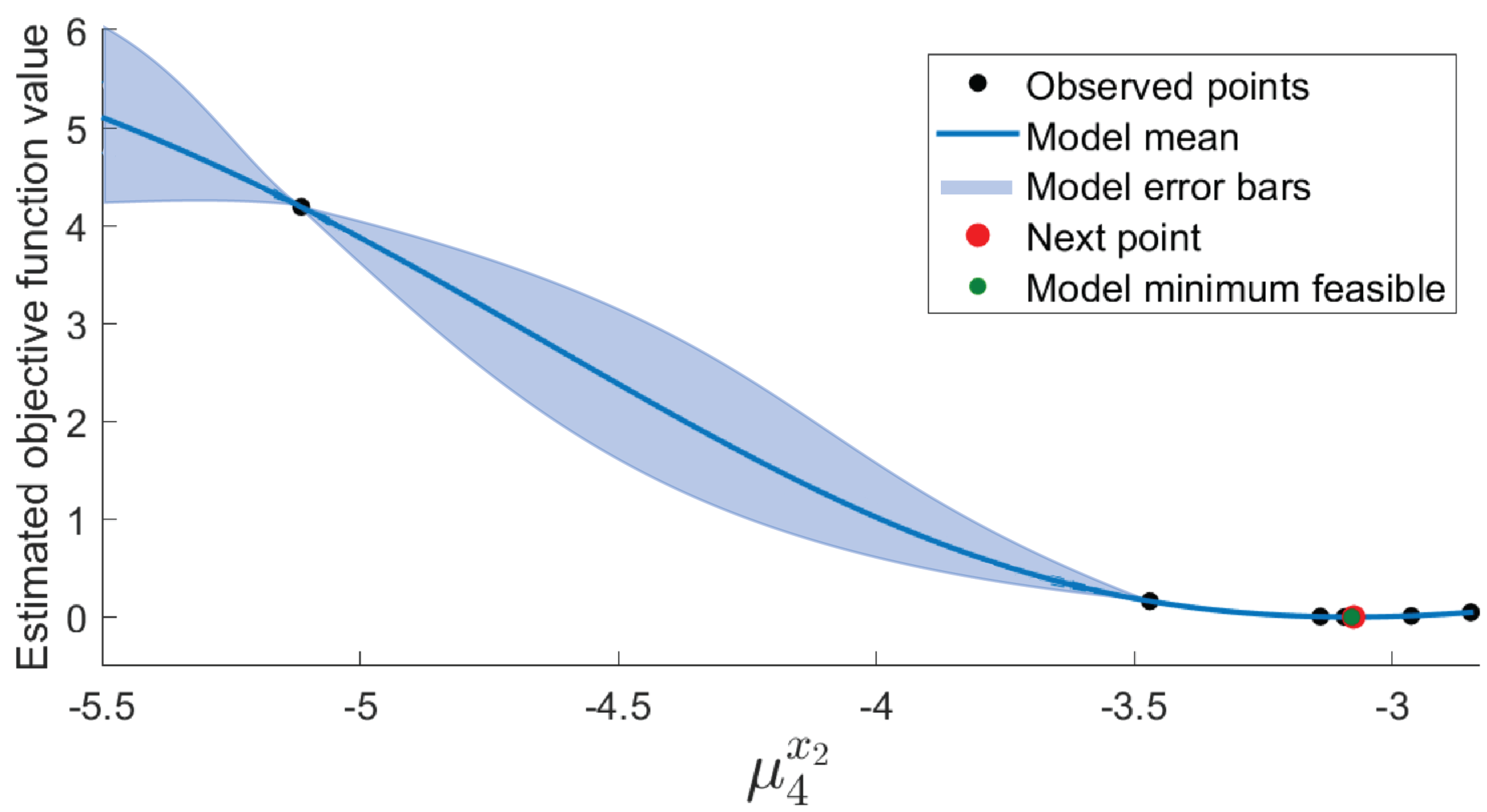}
	\includegraphics[width=.45\textwidth]{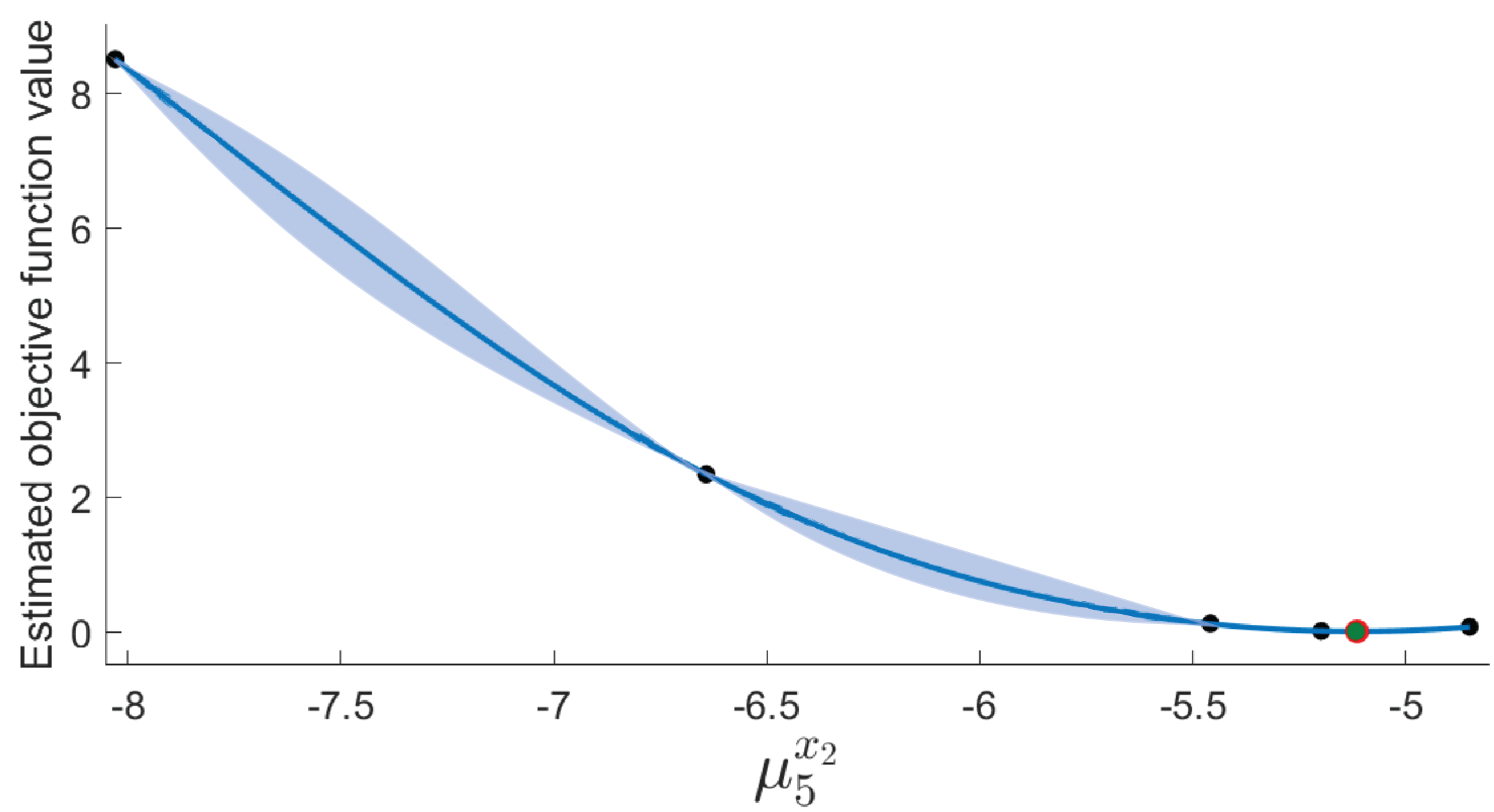}
	\caption{2D pick-and-place task (C-shape trajectories). The plots show the mean and confidence intervals estimated with a Gaussian process (GP) as surrogate model of BayesOpt after convergence (shaded area delimits $95\%$ credible intervals). The GPs correspond to the force-guided BayesOpt adaptation of HSMM states four (top) and five (bottom), depicted as purple and green ellipses in Fig. \ref{Fig:NomAdapPlan}. }
	\label{Fig:GPposterior}
\end{figure}

Note that the local adaptation of the HSMM states directly modifies the reference trajectory distribution that the robot tracks. The second and fourth plots in Fig.~\ref{Fig:NomAdapPlan} display the resulting reference distributions as blue solid lines (means) and shared areas (covariances) for both C- and L-shape models. As a result of the local adaptation, the retrieved trajectories match the nominal plan when no external forces were applied, and they deform according to the local changes imposed by the force-guided BayesOpt. More specifically, the new means $\bm{\mu}_i^{\mathcal{O}^*}$ and $\mu_i^{\mathcal{D}^*}$ computed by BayesOpt directly affect the adapted trajectory distribution computed via~\eqref{eq:trajHMM}. This local adaptation allows the user to introduce small changes in the trajectory without affecting relevant patterns of the task, such as the start and end of the reference distributions in Fig.~\ref{Fig:NomAdapPlan}, which are crucial for reaching and releasing the manipulated object. 

Figure~\ref{Fig:GPposterior} shows the Gaussian process posterior for the force-guided adaptation of the fourth (purple Gaussian) and fifth (green Gaussian) states of the C-shape model. Note that BayesOpt efficiently converged after only 6 and 5 iterations, respectively. The local adaptations leaded to disagreement forces close to zero, meaning that the new trajectory distribution was in accordance with the desired modifications imposed by the external force. We also ran our force-guided BayesOpt for the case in which the full set of means $\{\bm{\mu}_{s,i}, \mu^{\mathcal{D}}_{i}\}_{i=1}^K$ is considered, in contrast to confining the exploration through~\eqref{eq:argmax_i}. In this case, BayesOpt did not converge. We attribute this to the fact that the vector $\bm{\theta} \in \mathbb{R}^{18}$, leading to the curse of dimensionality, a well-known limitation of BayesOpt. Although dimensionality reduction techniques may be applied, we instead exploited the task model to locally restrict the optimization process so that the search space dimensionality was low.

%%%%%%%%%%%%%%%%%%%%%%%%%%%%%%%%%%%%%%%%%%%%%%%%%%%%%%%%%%%%%%%%%%%%%%%%%%%%%%%%
\section{DISCUSSION}
Note that the proposed trajectory adaptation has several aspects in common with a classical impedance control approach. Although we here assume that we can exploit the impedance control capabilities of the robot to physically interact with it, our approach differs from this classical control method in that the human provides force cues to modify the reference trajectory that the robot tracks, as in~\cite{Losey18:interactiveTrajDeformation}. This approach is advantageous as the human operator can experience the rendered compliance of the impedance controller, and deform a segment of the desired trajectory. Because of our local adaptation strategy, the deformed trajectory returns to the nominal plan when no interaction is detected, and so the robot can also contribute toward completing the shared task. Another advantage of the force-guided BayesOpt is the possibility to include constraints into the optimization problem which may aim at minimizing jerky motions, enforcing safe physical interaction, or restricting high deviations from the nominal plan. 

In our force-guided robot adaptation we omitted the covariance parameters for simplicity. Extending the vector or local model parameters~\ref{eq:thetaParams} to include the covariance terms is not trivial as $\bm{\Sigma}_i$ imposes positive-definiteness constraints for BayesOpt. In this line, two potential solutions are: \emph{(i)} to consider the Cholesky decomposition of $\bm{\Sigma}_i$ as proposed in~\cite{Abu-Dakka18:ForceImpedanceLearning} to learn variable stiffness matrices, or \emph{(ii)} to exploit the geometry of symmetric positive definite matrices as proposed in~\cite{Rozo17:Manipulability} to learn robot manipulability ellipsoids.

%%%%%%%%%%%%%%%%%%%%%%%%%%%%%%%%%%%%%%%%%%%%%%%%%%%%%%%%%%%%%%%%%%%%%%%%%%%%%%%%
\section{CONCLUSIONS AND FUTURE WORK}
\label{sec:Conclusions}
This paper introduced a data-efficient optimization framework aimed at adapting trajectory distributions as a function of interaction forces generated by a human operator. The proposed approach assumed that the force data produced by the human convey information regarding the human intended adaptation. These data were used to compute disagreement forces that the robot sought to minimize by locally adapting its task model, previously learned from demonstrations. Our approach leveraged the benefits of BayesOpt and the probabilistic encoding provided by HSMM to efficiently adapt spatio-temporal patterns in a handful of iterations. Smooth adapted trajectories were retrieved by taking advantage of the static and dynamic features encapsulated in the learned model in the form of a mixture of Gaussian distributions.
The reported experiments did not tackle the adaptation of the model covariance matrices as this entails a high-dimensional problem for BayesOpt. We plan to study how latent space representations of the HSMM parameters may be exploited to alleviate this issue. Moreover, we will evaluate the proposed framework in different real scenarios, considering tasks where time-varying force profiles are relevant for the successful performance.  

%%%%%%%%%%%%%%%%%%%%%%%%%%%%%%%%%%%%%%%%%%%%%%%%%%%%%%%%%%%%%%%%%%%%%%%%%%%%%%%%
%%%%%%%%%%%%%%%%%%%%%%%%%%%%%%%%%%%%%%%%%%%%%%%%%%%%%%%%%%%%%%%%%%%%%%%%%%%%%%%%
\addtolength{\textheight}{-12cm}   % This command serves to balance the column lengths
                                  % on the last page of the document manually. It shortens
                                  % the textheight of the last page by a suitable amount.
                                  % This command does not take effect until the next page
                                  % so it should come on the page before the last. Make
                                  % sure that you do not shorten the textheight too much.

\bibliographystyle{IEEEtran}
\bibliography{IEEEabrv,References_IROS19}

\begin{thebibliography}{10}
\providecommand{\url}[1]{#1}
\csname url@rmstyle\endcsname
\providecommand{\newblock}{\relax}
\providecommand{\bibinfo}[2]{#2}
\providecommand\BIBentrySTDinterwordspacing{\spaceskip=0pt\relax}
\providecommand\BIBentryALTinterwordstretchfactor{4}
\providecommand\BIBentryALTinterwordspacing{\spaceskip=\fontdimen2\font plus
\BIBentryALTinterwordstretchfactor\fontdimen3\font minus
  \fontdimen4\font\relax}
\providecommand\BIBforeignlanguage[2]{{%
\expandafter\ifx\csname l@#1\endcsname\relax
\typeout{** WARNING: IEEEtran.bst: No hyphenation pattern has been}%
\typeout{** loaded for the language `#1'. Using the pattern for}%
\typeout{** the default language instead.}%
\else
\language=\csname l@#1\endcsname
\fi
#2}}

\bibitem{ReedPeshkin:PhysicalHRI08}
K.~Reed and M.~Peshkin, ``Physical collaboration of human-human and human-robot
  teams,'' \emph{{IEEE} Transactions on Haptics}, vol.~1, no.~2, pp. 108--120,
  2008.

\bibitem{Nikolaidis17:Models4HRC}
S.~Nikolaidis, J.~Forlizzi, D.~Hsu, J.~Shah, and S.~Srinivasa, ``Mathematical
  models of adaptation in human-robot collaboration,'' \emph{ArXiv e-prints},
  July 2017.

\bibitem{Argall10:TactileCorrection}
B.~Argall, E.~L. Sauser, and A.~Billard, ``Tactile guidance for policy
  refinement and reuse,'' in \emph{Joint {IEEE} Intl. Conf. on Development and
  Learning and Epigenetic Robotics}, Ann Arbor, USA, August 2010, pp. 7--12.

\bibitem{Chu16:HapticAffordance}
V.~Chu, B.~Akgun, and A.~Thomaz, ``Learning haptic affordances from
  demonstration and human-guided exploration,'' in \emph{{IEEE} Haptics
  Symposium ({HAPTICS})}, Philadelphia, USA, April 2016, pp. 119--125.

\bibitem{Chu17:GuidedExploration}
V.~Chu and A.~Thomaz, ``Analyzing differences between teachers when learning
  object affordances via guided-exploration,'' \emph{Intl. Journal of Robotics
  Research}, vol.~36, no. 5--7, pp. 739--758, 2017.

\bibitem{Schroecker16:DirectingPolicySearch}
Y.~Schroecker, H.~B. Amor, and A.~Thomaz, ``Directing policy search with
  interactively taught via-points,'' in \emph{Intl. Conf. on Autonomous Agents
  and Multiagent Systems ({AAMAS})}, Singapore, May 2016, pp. 1052--1059.

\bibitem{Shahriari16:BayesOpt}
B.~Shahriari, K.~Swersky, Z.~Wang, R.~P. Adams, and N.~de~Freitas, ``Taking the
  human out of the loop: A review of {B}ayesian optimization,''
  \emph{Proceedings of the {IEEE}}, vol. 104, no.~1, pp. 148--175, 2016.

\bibitem{Cully15:RobotsAsAnimal}
A.~Cully, J.~Clune, D.~Tarapore, and J.~B. Mouret, ``Robots that can adapt like
  animals,'' \emph{Nature}, vol. 521, pp. 503--507, 2015.

\bibitem{Marco16:LQRbayesOpt}
A.~Marco, P.~Hennig, J.~Bohg, S.~Schaal, and S.~Trimpe, ``Automatic lqr tuning
  based on gaussian process global optimization,'' in \emph{{IEEE} Intl. Conf.
  on Robotics and Automation ({ICRA})}, Stockholm, Sweden, May 2016, pp.
  270--277.

\bibitem{Antonova17:DeepKernelsBO}
R.~Antonova, A.~Rai, and C.~Atkeson, ``Deep kernels for optimizing locomotion
  controllers,'' in \emph{Conference on Robot Learning ({CoRL})}, California,
  USA, November 2017, pp. 1--10.

\bibitem{Drieb17:ConstBayesOptForceTask}
D.~Drieß, P.~Englert, and M.~Toussaint, ``Constrained {B}ayesian optimization
  of combined interaction force/task space controllers for manipulations,'' in
  \emph{{IEEE} Intl. Conf. on Robotics and Automation ({ICRA})}, Singapore, May
  2017, pp. 902--907.

\bibitem{Ghadirzadeh16:RL4pHRI}
A.~Ghadirzadeh, J.~B\"{u}tepage, A.~Maki, D.~Kragic, and M.~Bj\"{o}rkman, ``A
  sensorimotor reinforcement learning framework for physical human-robot
  interaction,'' in \emph{{IEEE/RSJ} Intl. Conf. on Intelligent Robots and
  Systems ({IROS})}, Daejeon, Korea, October 2016, pp. 2682--2688.

\bibitem{Kupcsik15:HandoverBayes}
A.~Kupcsik, D.~Hsu, and S.~Lee, ``Learning dynamic robot-to-human object
  handover from human feedback,'' in \emph{Intl. Symp. on Robotics Research},
  Sestri Levante, Italy, September 2015.

\bibitem{Yu:HSMM10}
S.~Yu, ``Hidden semi-{M}arkov models,'' \emph{Artificial Intelligence}, vol.
  174, no.~2, pp. 215--243, 2010.

\bibitem{Rozo16Front:ADHSMM}
L.~Rozo, J.~Silv\'erio, S.~Calinon, and D.~G. Caldwell, ``Learning controllers
  for reactive and proactive behaviors in human-robot collaboration,''
  \emph{Frontiers in Robotics and {AI}}, vol.~3, no.~30, pp. 1--11, June 2016,
  specialty Section Robotic Control Systems.

\bibitem{Losey18:interactiveTrajDeformation}
S.~Losey and M.~Malley, ``Trajectory deformations from physical human--robot
  interaction,'' \emph{{IEEE} Transactions on Robotics}, vol.~34, no.~1, pp.
  126--138, 2018.

\bibitem{Berkenkamp16:safeBO}
F.~Berkenkamp, A.~Schoellig, and A.~Krause, ``Safe controller optimization for
  quadrotors with {G}aussian processes,'' in \emph{{IEEE} Intl. Conf. on
  Robotics and Automation ({ICRA})}, Stockholm, Sweden, May 2016, pp. 491--496.

\bibitem{RozoEtAl11:HMMGMRa}
L.~Rozo, P.~Jim\'{e}nez, and C.~Torras, ``Robot learning from demonstration of
  force-based tasks with multiple solution trajectories,'' in \emph{{IEEE}
  Intl. Conf. on Advanced Robotics ({ICAR})}, Tallin, Estonia, June 2011, pp.
  124--129.

\bibitem{RozoEtAl13:ForceLfD}
------, ``A robot learning from demonstration framework to perform force-based
  manipulation tasks,'' \emph{Journal of Intelligent Service Robotics, Special
  Issue on Artificial Intelligence Techniques for Robotics: Sensing,
  Representation and Action, Part 2}, vol.~6, no.~1, pp. 33--51, 2013.

\bibitem{Tanwani19:nonParamRobotLearn}
A.~K. Tanwani and S.~Calinon, ``Small-variance asymptotics for non-parametric
  online robot learning,'' \emph{Intl. Journal of Robotics Research}, vol.~38,
  no.~1, pp. 3--22, 2019.

\bibitem{MedinaEtAl:CollForce11}
J.~Medina, M.~Lawitzky, A.~Mortl, D.~Lee, and S.~Hirche, ``An experience-driven
  robotic assistant acquiring human knowledge to improve haptic cooperation,''
  in \emph{{IEEE/RSJ} Intl. Conf. on Intelligent Robots and Systems ({IROS})},
  San Francisco, USA, September 2011, pp. 2416--2422.

\bibitem{Rabiner89:TutorialHMM}
L.~Rabiner, ``A tutorial on hidden {M}arkov models and selected applications in
  speech recognition,'' in \emph{Proceedings of the {IEEE}}, 1989, pp.
  257--286.

\bibitem{YuKobayashi:efficientHSMM06}
S.~Yu and T.~Kobayashi, ``Practical implementation of an efficient
  forward-backward algorithm for an explicit-duration hidden {M}arkov model,''
  \emph{{IEEE} Trans. on Signal Processing}, vol.~54, no.~5, pp. 1947--1951,
  2006.

\bibitem{SugiuraEtal:MotionLearningTrajHMM11}
K.~Sugiura, N.~Iwahashi, H.~Kashioka, and S.~Nakamura, ``Learning, generation,
  and recognition of motions by reference-point-dependent probabilistic
  models,'' \emph{Advanced Robotics}, vol.~25, no. 6--7, pp. 825--848, 2011.

\bibitem{Mockus75:EI}
J.~Mo{\v{c}}kus, ``On {B}ayesian methods for seeking the extremum,'' in
  \emph{Optimization Techniques IFIP Technical Conference}, 1975, pp. 400--404.

\bibitem{Yin18:HumanInRobotMotionLearning}
H.~Yin, ``Incorporating human expertise in robot motion learning and
  synthesis,'' Ph.D. dissertation, \'{E}cole Polytechnique F\'{e}d\'{e}rale De
  Laussane, 2018.

\bibitem{Abu-Dakka18:ForceImpedanceLearning}
F.~J. Abu-Dakka, L.~Rozo, and D.~G. Caldwell, ``Force-based variable impedance
  learning for robotic manipulation,'' \emph{Robotics and Autonomous Systems},
  vol. 109, pp. 156--167, 2018.

\bibitem{Rozo17:Manipulability}
L.~Rozo, N.~Jaquier, S.~Calinon, and D.~G. Caldwell, ``Learning manipulability
  ellipsoids for task compatibility in robot manipulation,'' in
  \emph{{IEEE/RSJ} Intl. Conf. on Intelligent Robots and Systems ({IROS})},
  Vancouver, Canada, September 2017, pp. 3183--3189.

\end{thebibliography}

\end{document}